\documentclass[sigconf]{acmart}

\AtBeginDocument{%
  \providecommand\BibTeX{{%
    \normalfont B\kern-0.5em{\scshape i\kern-0.25em b}\kern-0.8em\TeX}}}

\usepackage{xcolor,colortbl,soul}
\usepackage{subcaption}
\usepackage{graphicx}
\usepackage{hhline}
\usepackage{multirow}
\usepackage{siunitx}
\usepackage{hyperref}
\usepackage{xcolor,colortbl}
\usepackage{balance}
\usepackage{caption}
\usepackage{enumitem}
\usepackage{listings}

\definecolor{Gray}{gray}{0.85}

\setcopyright{none}
\renewcommand\footnotetextcopyrightpermission[1]{} 
\begin{document}

\title{A Convolutional LSTM based Residual Network for Deepfake Video Detection}




\author{Shahroz Tariq, Sangyup Lee}
\orcid{1234-5678-9012}
\affiliation{%
  \institution{Dept. of Computer Science and Engineering\\ Sungkyunkwan University}
  \city{Suwon} 
  \state{South Korea} 
}
\email{{shahroz,sangyup.lee}@g.skku.edu}


\author{Simon S. Woo}
\authornote{Corresponding Author}
\affiliation{%
  \institution{Dept. of Applied Data Science\\ Sungkyunkwan University}
  \city{Suwon} 
  \country{South Korea}
  }
\email{swoo@g.skku.edu }



\begin{abstract}
In recent years, deep learning-based video manipulation methods have become widely accessible to masses. With little to no effort, people can easily learn how to generate deepfake videos with only a few victims or target images. This creates a significant social problem for everyone whose photos are publicly available on the Internet, especially on social media websites. Several deep learning-based detection methods have been developed to identify these deepfakes. However, these methods lack generalizability, because they perform well only for a specific type of deepfake method. Therefore, those methods are not transferable to detect other deepfake methods. Also, they do not take advantage of the temporal information of the video. In this paper, we addressed these limitations. We developed a Convolutional LSTM based Residual Network (CLRNet), which takes a sequence of consecutive images as an input from a video to learn the temporal information that helps in detecting unnatural looking artifacts that are present between frames of deepfake videos. We also propose a transfer learning-based approach to generalize different deepfake methods. Through rigorous experimentations using the FaceForensics++ dataset, we showed that our method outperforms five of the previously proposed state-of-the-art deepfake detection methods by better generalizing at detecting different deepfake methods using the same model. 
\end{abstract}



\keywords{Deepfake detection, Video forensics, Image manipulation}


\maketitle

\section{Introduction}
\label{sec:intro}

Deep learning-based methods for synthetic image generation have sprouted tremendously in the last few years. These new methods can generate photorealistic images that can easily deceive average humans~\cite{FaceForensics++,Face2Face,FaceSwap,Deepfakes,NeuralTextures,PGGAN}. Due to their ability, these methods have many applications in computer vision or graphics disciplines, such as human face generation~\cite{PGGAN} and photorealistic scenery generation~\cite{GauGAN}. However, there is also a dark side to all of this innovation. Many people with malicious intentions have used these methods to generate fake videos of celebrities and masses~\cite{news1,news2,news3,news4}, for which numerous approaches exist~\cite{Face2Face,FaceSwap,NeuralTextures,Deepfakes}. This has started causing major social issues: a recent study claimed that 96\% of the deepfakes originate from porn videos~\cite{news5}. They come under the same umbrella of so-called Deepfakes.
Recently, the research community has released numerous deepfake datasets to assist other researchers in developing detection mechanisms for these deepfakes. The most pioneering work is the FaceForensics++ dataset~\cite{FaceForensics++} developed in part by Google. Originally, the FaceForensics++~\cite{FaceForensics++} dataset contained Pristine (1,000), Deepfakes (1,000), FaceSwap (1,000), Face2Face (1,000), and Neural Texture (1,000) videos. Later, Google contributed by supplementing real (363) and fake (3,000) videos. This year, Facebook launched a deepfake detection challenge with prize money of one million U.S. dollars to accelerate research in this field~\cite{DFDCWebsite}.
Lately, several deepfake detection methods with high zero-shot test accuracy on specific training deepfake datasets have emerged~\cite{ForensicTransfer,DFD1,DFD3,Shahroz2,Hyeonseong1,Hyeonseong2}. However, they would have poor detection accuracy on new deepfake methods that were not present in the training set. This was our primary motivation to develop a generic and universal deepfake video detector, since it would be impractical to produce datasets for every new deepfake generation method. Therefore, 
we leverage massive deepfake datasets, such as FaceForensics++~\cite{FaceForensics++}, that are already available and employ transfer learning to detect other deepfake methods as well as newly generated ones.

Studies on the development of a generic deepfake detector have seen limited research activity~\cite{ForensicTransfer}. Therefore, we aim to address these problems by developing a model that first trains on one deepfake method from the massive deepfake dataset of FaceForensics++~\cite{FaceForensics++} and then uses a few-shot transfer learning method to learn about other deepfake methods. The number of sample videos required for few-shot learning is minimal, rendering the choice more practical. Our method differs from previous works in that we explored different transfer learning strategies and compared their results through rigorous experimentations on multiple datasets. 

\begin{figure}[t]
    \centering
    \includegraphics[width=\columnwidth]{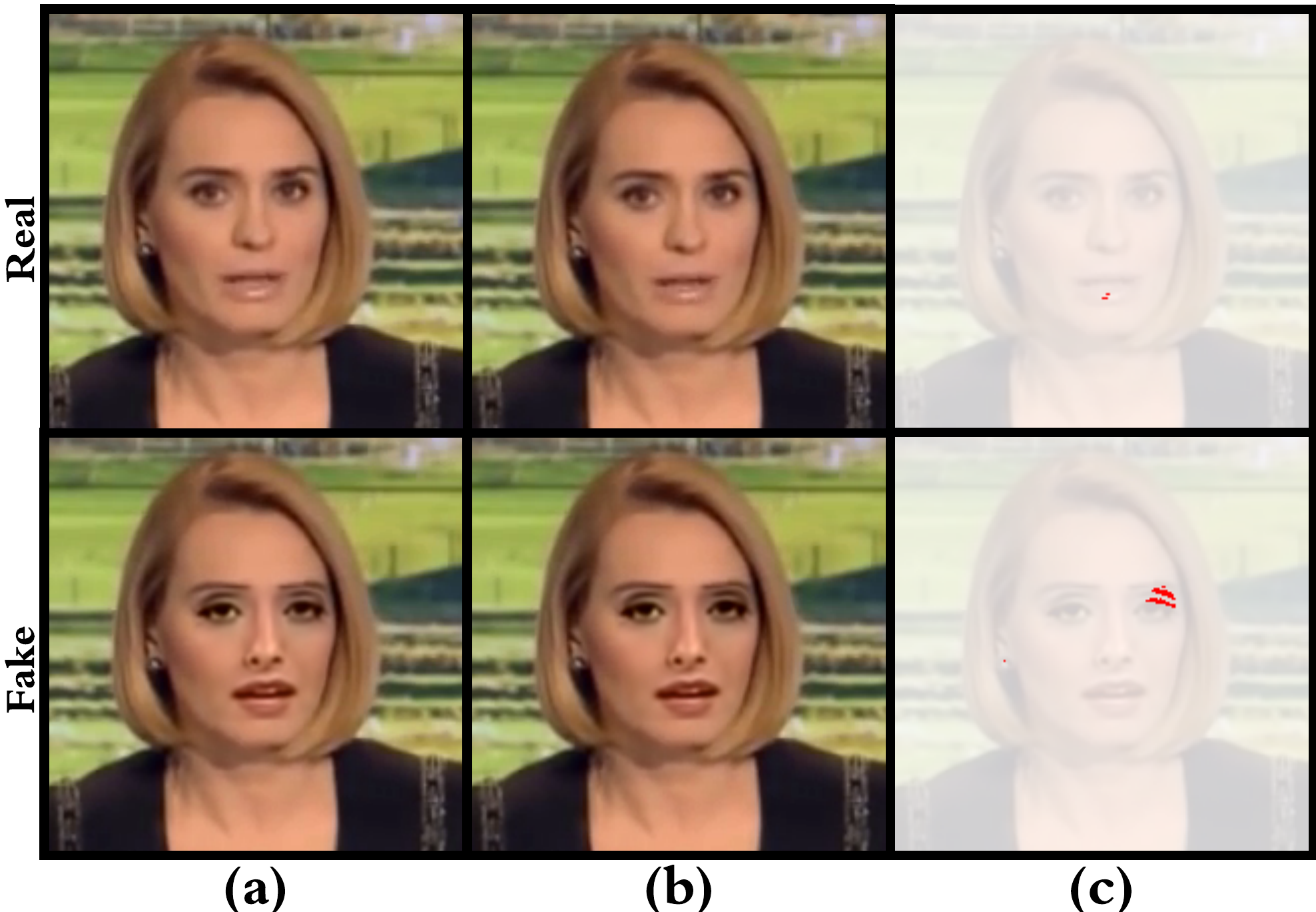}
    \caption{Difference between two consecutive frames of a deepfake video. (a) and (b) are the $n^{th}$ and $(n+1)^{th}$ frames, respectively. For pristine videos, these frames are nearly the same, but for deepfake videos, there are inconsistencies. The difference between the $n^{th}$ and $(n+1)^{th}$ frames highlights (red) those inconsistencies in (c). There is a small difference between real video frames, but the difference between fake video frames is more significant, as shown in (c).}
    \label{fig:inconsistency}
\end{figure}

We noticed that most of the deepfake detection methods~\cite{ForensicTransfer,FaceForensics++} randomly extract frames (images) from videos for training and testing, hence a single frame-based detection method. However, after carefully observing numerous deepfake videos, we were surprised to discover tiny artifacts between consecutive frames within the deepfake videos through which we can identify these videos, as shown in Fig.~\ref{fig:inconsistency}. Therefore, we concluded that the temporal information between consecutive video frames is crucial for deepfake detection~\cite{DFD2}. To incorporate the temporal aspect, we used a Convolutional LSTM, since it has shown to be useful for such tasks~\cite{ConvLSTM1}. Therefore, we propose CLRNet, a Convolutional LSTM based Residual Network for Deep Fake Video Detection using Transfer Learning.

The main contributions are summarized as follows:

\begin{itemize}
    \item \textbf{CLRNet}: We propose a novel architecture based on Convolutional LSTM and Residual Network for deepfake detection using a sequence of consecutive frames from a video.
    \item \textbf{Generalizability}: We provided a more generalized method than previous state-of-the-art deepfake detection approaches with high accuracy and demonstrated it through rigorous experimentations.
\end{itemize}

\section{Related Work}
\label{sec:related}

Our work spans across different fields, such as deepfake detection, model generalization, and transfer learning. Therefore, we will briefly cover the related works in this section.

\subsection{Deepfake Detection}The detection of abnormal eye blinking~\cite{eyeblink} has shown to be effective for the identification of inconsistencies in manipulated videos or images. Furthermore, image splice detection methods~\cite{Splice1,FT36,FT6} aim to exploit the deviation resulting from splicing near the boundaries of manipulated regions in an image. Although inconsistencies in images generated from existing deepfake generation methods can be detected, new and more advanced generation methods are researched and developed every year. On the other hand, model-based methods, such as the measurement of features from demosaicing artifacts~\cite{FT16}, lens aberrations~\cite{FT41}, and JPEG artifacts resulting from the choice of different image processing methods~\cite{FT4}, have traditionally been used to identify image manipulations. Nevertheless, they have shown to be unreliable when detecting machine-generated fake images, such as Generative Adversarial Networks (GANs) and Variational Autoencoders (VAEs), because the entire set of images are created from scratch. Deep learning-based approaches in a supervised environment have shown high detection accuracy. Specifically, Convolutional Neural Network (CNN) based approaches concentrated on automatically learning hierarchical representations from the RGB color images input~\cite{FT27,FT33} or utilizing manipulation detection features~\cite{FT7}, and using hand-crafted features~\cite{FT11}. Tariq \textit{et al.}~\cite{Shahroz1,Shahroz2} introduced ShallowNet, a fast learning and effective CNN-based network for detecting GAN-generated images with high accuracy even at low resolution (64$\times$64). Furthermore, Zhou \textit{et al.}~\cite{FT44} applied a two-stream Faster R-CNN network, which can capture high and low-level image details. R\"ossler \textit{et al.}~\cite{FaceForensics,FaceForensics++} presented a significantly improved performance on compressed images, which is essential for detecting deepfakes on social networking sites such as Instagram, Facebook, and Twitter. Most of the aforementioned approaches concentrate on detecting facial manipulations in a single video frame. However, as shown in Fig.~\ref{fig:inconsistency}, it is crucial to analyze the temporal information between consecutive frames in deepfake videos. In our approach, we use multiple consecutive frames to utilize this temporal information for an improved detection of deepfakes.

\subsection{Detection with Consecutive Video Frames}Sabir \textit{et al.}~\cite{DFD2} proposed a detection method that utilizes both the CNN and Recurrent Neural Network (RNN) to capture the temporal information presented in 5 consecutive deepfake video frames. Also, G\"uera \textit{et al.}~\cite{FT_SQ2} adopted a similar approach, extracting the features from up to 80 consecutive frames using CNN layers and feeding them to RNN layers to build a temporal information-aware deepfake detection model. Both methods~\cite{DFD2,FT_SQ2} extract features from CNN and pass it to RNN layers. At the same time, we build our CLRNet model using Convolutional LSTM cells, which can capture the spatio-temporal information directly from an input image sequence.
However, most of these approaches yielded worse results when evaluated on datasets containing videos from a different deepfake generation method. Thus, we explored transfer learning for our CLRNet model to address this challenge.

\begin{figure}[t!]
    \centering
    \includegraphics[width=\columnwidth]{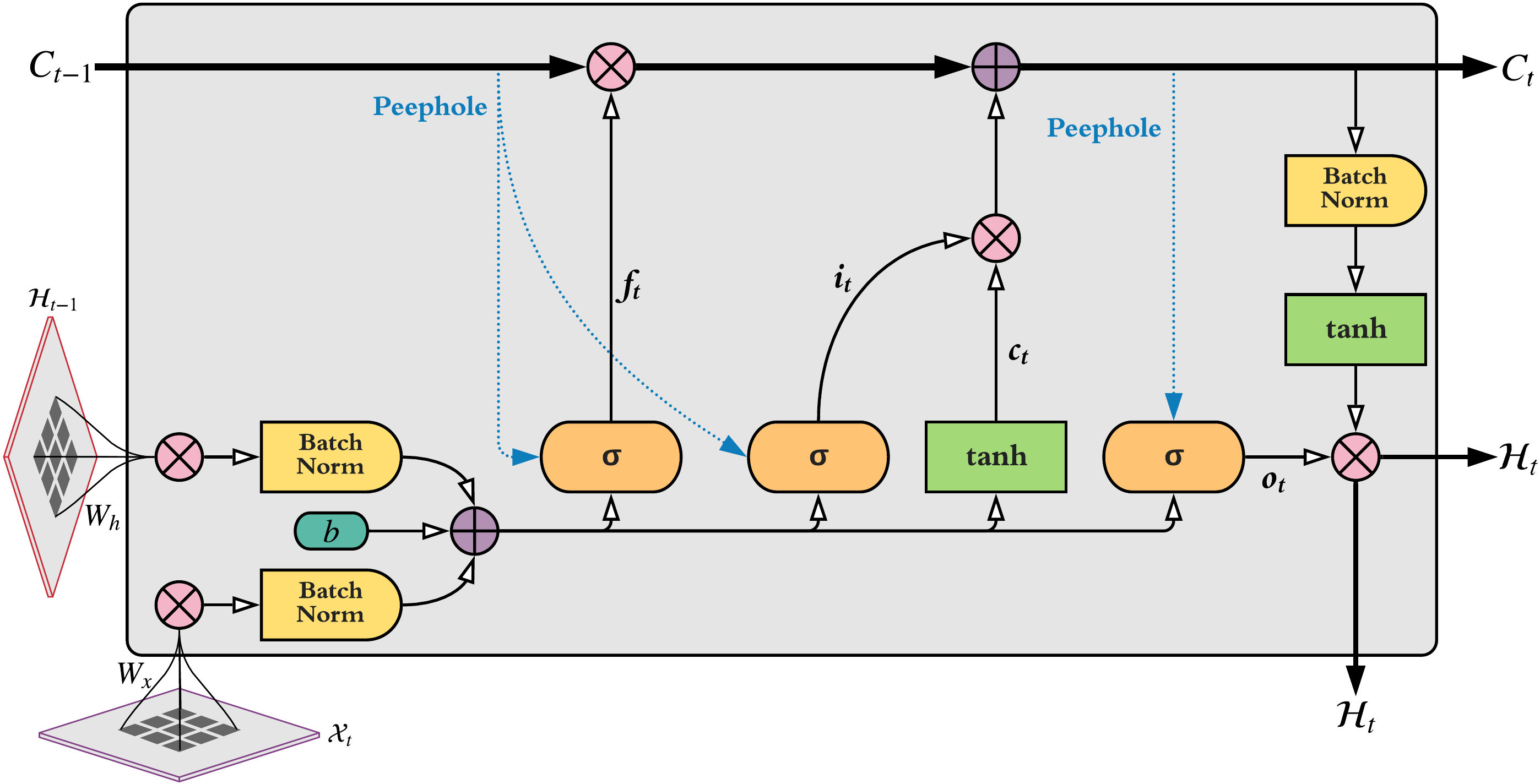}
    \caption{\emph{Convolutional LSTM cell}: Visual representation of the Convolutional LSTM cell. The main structure is similar to that of an LSTM, but some additional components are present to incorporate the convolutional part. Here, $\mathcal{X}_t$ is the input, $C_t$ is the cell output, and $\mathcal{H}_t$ is the hidden state. The gates are represented by $i_t$, $f_t$, and $o_t$.}
    \label{fig:convlstm_cell}
\end{figure}

\subsection{Generalization via Transfer Learning}A variety of deepfake video generation techniques are constantly being developed and more sophisticated deep fake videos will arise in the future. However, collecting and producing a significant amount of new deepfake samples would be impractical. To cope with such situations, few-shot transfer learning (TL) is the key to the detection of deepfakes created by different methods. That is, what has been learned in one domain (e.g., FaceSwap) can be used to enhance the generalizability in another domain (e.g., Face2Face). 
Cozzolino \textit{et al.}~\cite{ForensicTransfer} have experimented the generalization of a single detection method for multiple target domains. In this work, we compare our approach against ForensicsTransfer~\cite{ForensicTransfer} to demonstrate the enhanced generalizability and transferability. 


\begin{figure}[t!]
    \centering
    \includegraphics[width=\linewidth]{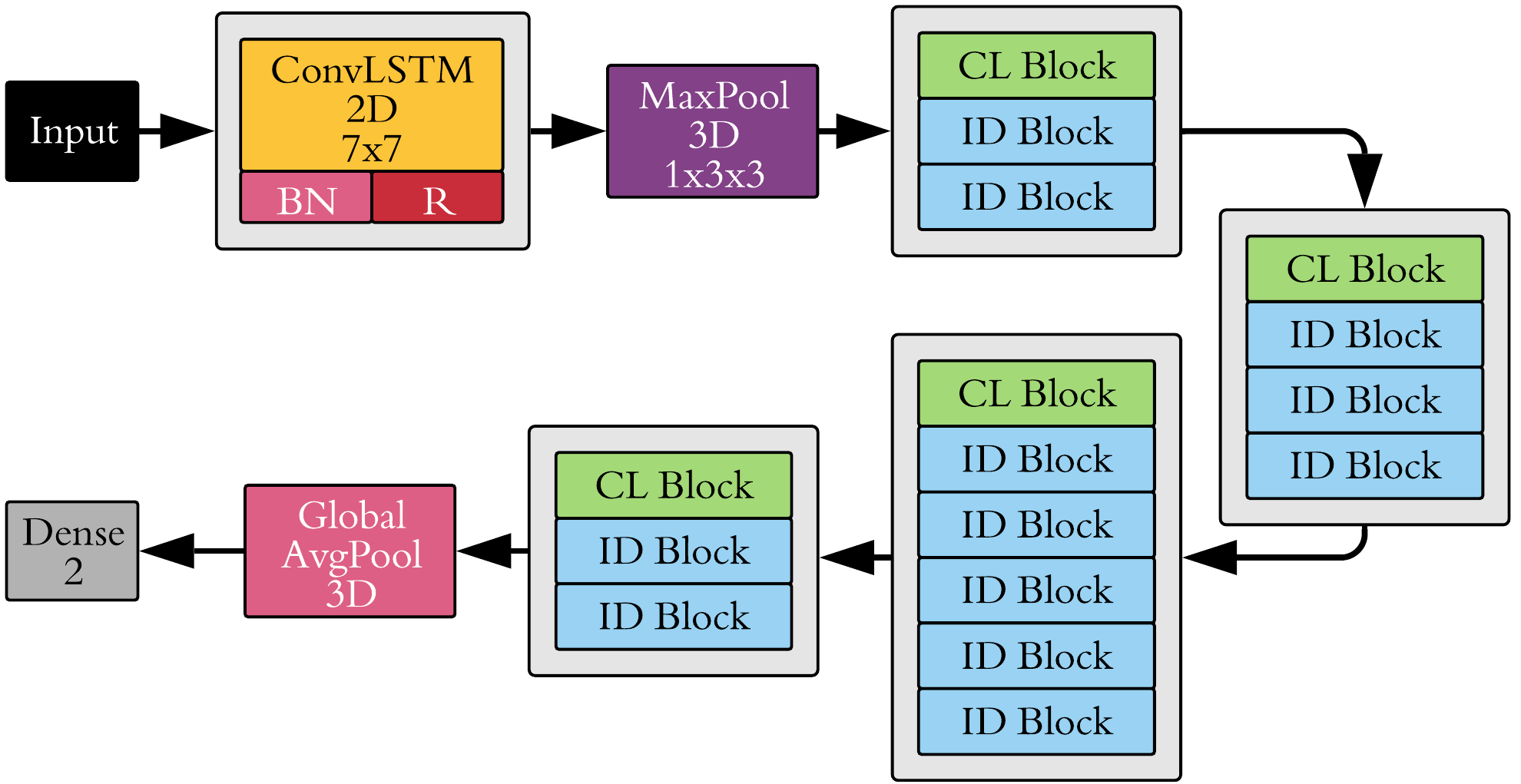}
    \caption{\emph{CLRNet Architecture}: High-level architectural diagram of our Convolutional LSTM based Residual Network (CLRNet) model. The input to the model is a sequence of consecutive images and the output is a classification result, real or fake. We used Keras temrs to denote the layer names.}
    \label{fig:CLRNet_model}
\end{figure}

\section{Our Approach}
\label{sec:approach}
A frame-by-frame analysis of deepfake videos reveals the inconsistencies between consecutive frames in deepfake videos, which are absent in pristine videos. These inconsistencies include 1) a sudden change in brightness and contrast on a small region of the face, and 2) the size of some facial parts such as eyes, lips, and eyebrows changes between frames. These are minor inconsistencies that can be detected with a thorough examination. Figure~\ref{fig:inconsistency} shows an example of such artifact from two consecutive frames of a deepfake video along with their sudden differences marked in red. These inconsistencies render the video somewhat unnatural. Motivated by this finding and observation, we developed a new deepfake detection method, the Convolutional LSTM Residual Network, that can account for these inconsistencies for the identification of real and fake videos.
\subsection{Convolutional LSTM Cell}
Shi \textit{et al.}~\cite{ConvLSTM1} stated that the main problem with handling spatio-temporal data in FC-LSTM~\cite{FC-LSTM} is the use of full connections during input-to-state and state-to-state transitions, and no spatial information is encoded. In contrast, Convolutional LSTM (ConvLSTM) overcomes this problem by introducing 3D tensors whose last two dimensions are spatial (rows and columns) for all the inputs $\left (\mathcal{X}_1, \dots ,\mathcal{X}_t\right)$, outputs $\left (C_1, \dots ,C_t\right)$, hidden states $\left (\mathcal{H}_1, \dots ,\mathcal{H}_t\right)$, and gates $\left (i_t,f_t,o_t\right)$. In this paper, we follow the formulation of ConvLSTM by Shi \textit{et al.}~\cite{ConvLSTM1}. The Hadamard product and the Convolution operator are denoted by `$\circ$' and `$\ast$', respectively. 
\begin{equation}
\small
    \begin{split}
        \label{eq:i}
        i_t &= \sigma \left( W_{x_i} \ast \mathcal{X}_{t} + W_{h_i} \ast \mathcal{H}_{t-1}+W_{c_i} \circ C_{t-1}+b_i \right)
        \\
        f_t &= \sigma \left( W_{x_f} \ast \mathcal{X}_{t} + W_{h_f} \ast \mathcal{H}_{t-1}+W_{c_f} \circ C_{t-1}+b_f \right)
        \\
        C_t &= f_t \circ C_{t-1} + i_t \circ \tanh \left (W_{x_c} \ast \mathcal{X}_{t}+W_{h_c} \ast \mathcal{H}_{t-1}+b_c \right)
        \\
        o_t &= \sigma \left( W_{x_o} \ast \mathcal{X}_t + W_{h_o} \ast \mathcal{H}_{t-1} + W_{c_o}\circ C_t + b_o \right)
        \\
        \mathcal{H}_t &= o_t \circ \tanh \left(C_t \right)
    \end{split}
\end{equation}
Furthermore, a visual representation of our ConvLSTM cell, based on Xavier~\cite{Convlstmcell} and implementation of Keras~\cite{Keras}, is shown in Fig.~\ref{fig:convlstm_cell}.
\begin{figure}[t!]
    \centering
    \includegraphics[width=1\columnwidth]{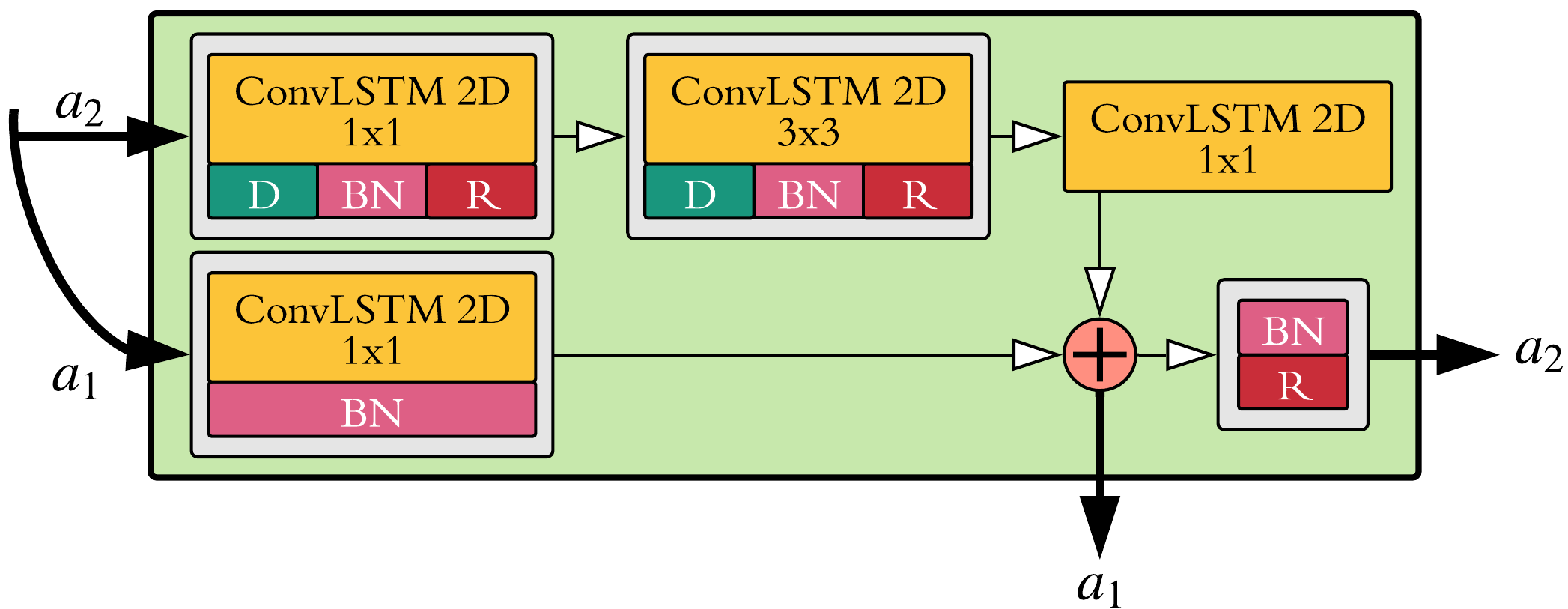}
    \caption{\emph{CL Block}: Visual representation of the internal structure of the Convolutional LSTM block (CL Block). There are two ConvLSTM2D layers, each followed by dropout, BatchNorm, and ReLU. Afterward, we directly added the last ConvLSTM2D layer to a ConvLSTM2D with a BatchNorm layer from the shortcut connection to get $a_1$. Finally, $a_2$ results from BatchNorm followed by ReLU applied to the output of the addition operator. The input $a_1$ and $a_2$ are identical for the first CL Block, but different afterward.}
    \label{fig:CL_block}
\end{figure}
\subsection{Convolutional LSTM Residual Network}
\begin{figure}[t!]
    \centering
    \includegraphics[width=1\columnwidth]{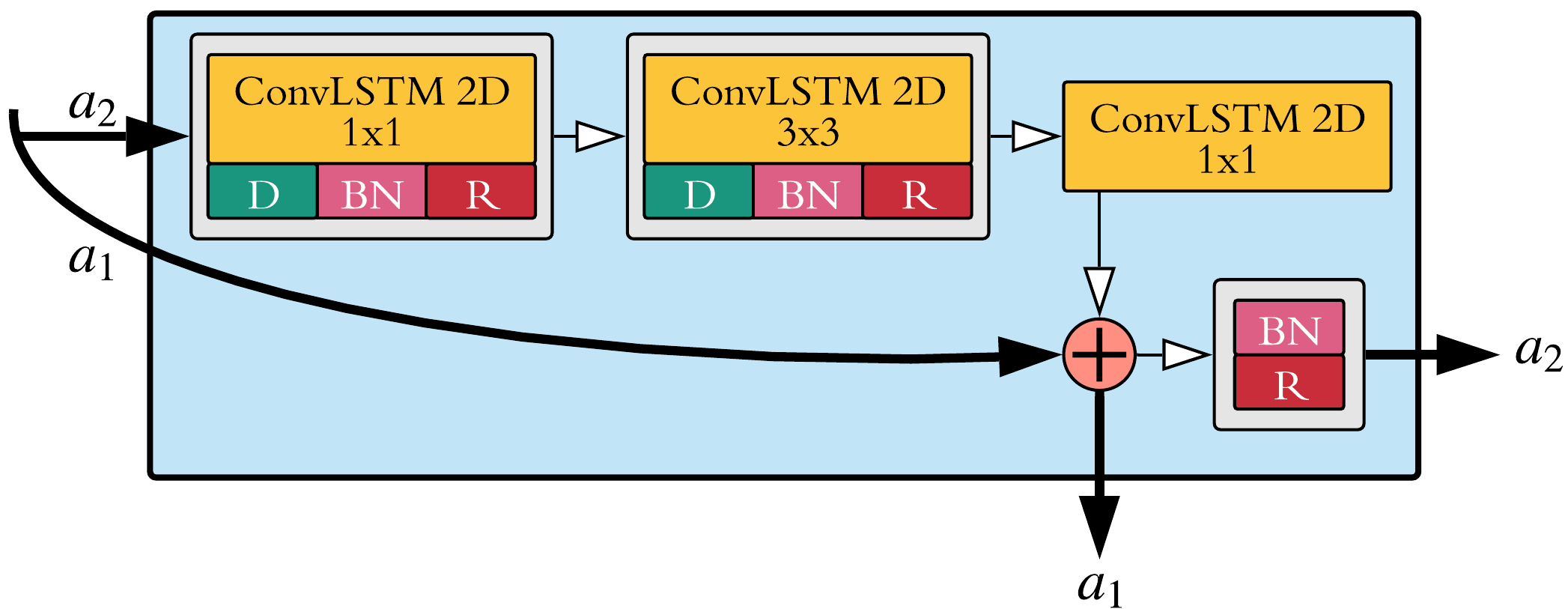}
    \caption{\emph{ID Block}: Visual representation of the internal structure of the Identity block (ID Block). Similar to the CL Block, there are two ConvLSTM2D layers, each followed by dropout, BatchNorm, and ReLU. Afterward, we add the last ConvLSTM2D layer to the input $a_1$ from the shortcut connection to get output $a_1$. Finally, $a_2$ is the result of BatchNorm and ReLU applied to the output of the addition operator.}
    \label{fig:ID_block}
\end{figure}

Sabir~\textit{et al.}~\cite{DFD2} showed that a CNN based backbone encoding network, such as ResNet or DenseNet, connected to an RNN based network achieves high accuracy for deepfake detection tasks. They also find a sequence of images for an improved performance compared to using a single frame input. Furthermore, to avoid the vanishing gradient problem, we added residuals in our network. Based on previous research~\cite{DFD2,DFD3,DFD6,DFD9} and our analysis of inconsistencies and/or artifacts in consecutive frames, we developed our Convolutional LSTM based Residual Network (CLRNet). Figure~\ref{fig:CLRNet_model} shows a visual representation of the architecture for our CLRNet model. The input elements of our model are 3D tensors preserving the entire spatial information for consecutive frames. Therefore, we used ConvLSTM (CL) cells instead of Convolution cells,as shown in Fig.~\ref{fig:CL_block} and ~\ref{fig:ID_block}. Shi \textit{et al.}~\cite{ConvLSTM1} state that stacking ConvLSTM in this way provides a strong representational power to the model.
We developed the core architecture of our CLRNet using two types of building blocks (i.e., CL block and ID block). Figures~\ref{fig:CL_block} and~\ref{fig:ID_block} provide a pictorial representation of the CL and ID blocks. These blocks can be related to the Convolution and Identity building blocks from ResNet~\cite{resnet}, respectively. The building blocks in our CLRNet model have two outputs (i.e., $a_1$ and $a_2$), as shown in Fig.~\ref{fig:CL_block} and~\ref{fig:ID_block}. In both blocks, $a_1$ is the output following the addition step, whereas $a_2$ is the output of the addition step followed by the batch normalization and ReLU layers. The output $a_1$ and $a_2$ serve as inputs for the next block. The CL block contains a ConvLSTM cell followed by a batch normalization layer on the shortcut path, whereas in the ID block, the shortcut path directly connects the input $a_1$ to the addition layer.

\subsection{Transfer Learning Strategies}
For transfer learning, we evaluated the following three strategies: 1) Single-source to Single-target: we train our model with one large deepfake dataset and then apply transfer learning to one target deepfake dataset (e.g., the model is first trained on the DF dataset and then transfer learned to FS), 2) Multi-source to Single-target: we train our model on multiple sources and apply transfer learning to a single target domain, and 3) Single-source to Multi-target: we train our model on a single source domain and then use a small volume of multiple target domains to apply transfer learning. We will discuss the advantages and disadvantages of each strategy in Section~\ref{sec:res}.

\subsection{Implementation Details}

\subsubsection{Dataset description.}
To compare our method with different baselines, we used DeepFake (DF), FaceSwap (FS), Face2Face (FS), NeuralTextures(NT), and DeepFakeDetection (DFD) datasets in~\cite{FaceForensics++}.
Table~\ref{tab:dataset_details} describes all the datasets used for this paper. Each class inside the FaceForensics++~\cite{FaceForensics++} dataset, except DFD, contains 1,000 videos. We used the first 750 videos out of 1,000 for training, the next 125 for validation, and the remaining of the 125 for testing. For DFD, we selected only 300 videos (250 for training, 25 for validation, and 25 for testing).

\subsubsection{Preprocessing.}
From each real and fake video, we extracted 16 samples such that each sample contains five consecutive frames. We used multi-task CNN (MTCNN)~\cite{mtcnn} to detect the face landmark information inside the extracted frame. Afterward, we used this landmark information to crop the face from the image and aligned it to the center. Lastly, we resized all the frames to a $240\times240$ resolution.

\subsubsection{Data Augmentation.}
We also applied data augmentation techniques to diversify the training data. We varied the following conditions: 1) Brightness (-30\% to 30\%), 2) Channel shift (-50 to 50), 3) Zoom (-20\% to 20\%), 4) Rotation (-30$^{\circ}$ degrees to 30$^{\circ}$), and 5) Horizontal flip (50\% probability).

\subsubsection{Training.}
The idea behind our method is to train on one of the widely available deepfake datasets and then use transfer learning to train for the other datasets with a small amount of samples. We trained our model by taking 16 samples from each of the 750 real (Pristine) and 750 fake (DF or FS or F2F or NT) videos.

\subsubsection{Transfer Learning Configuration.}
Once the training is complete for the base model, we performed transfer learning to other datasets by using a small subset of the target dataset (10 videos). For our CLRNet model, we freezed the first 120 layers and applied transfer learning to the model with an equal number of videos from the source and target datasets, that is, 10 videos per dataset.

\subsubsection{Machine Configuration.}
We used Intel(R) Xeon(R) Silver 4114 CPU @ 2.20GHz with 256.0GB RAM and NVIDIA GeForce Titan RTX. We used TensorFlow v1.13.0~\cite{tensorflow} with Keras Library~\cite{Keras} on Python v3.7.5 for the implementation of our CLRNet model. 

\subsubsection{Evaluation Metrics.}
We used Precision, Recall, and F1-Score for the evaluation. Due to space limitations, we are reporting only the F1-Scores in Table~\ref{tab:training} and \ref{tab:TL}. ShallowNet, Xception, FF++~\cite{FaceForensics++}, and FT uses a single frame as input for training and testing sets, whereas Sabir\textit{ et al.}~\cite{DFD2} and our CLRNet uses five consecutive frames as input. We kept the same number of real and fake images in the training, validation, and test sets to minimize the influence of data imbalance during evaluation.

\subsection{Baseline Methods}
We compared CLRNet with several state-of-the-art methods and tried our best to implement them according to their specifications. The following is a description of these methods.

\subsubsection{Xception.} The Xception Network~\cite{Xception} is considered as the state-of-the-art for image classification task. We used the Keras~\cite{Keras} implementation of Xception, which is pre-trained on the ImageNet dataset~\cite{ImageNet}.
\begin{table}
\centering
\caption{\emph{Dataset details}: There are 1,000 videos for Pristine, DeepFake, FaceSwap, Face2Face, and Neural Textures, respectively: we used 750 real and 750 fake videos for training, 125 real and 125 fake videos for validation as well as for testing. There are 3,363 videos (363 real, 3,000 fake) in the DeepFakeDetection dataset: we used 250 real and 250 fake videos for training, 25 real and 25 fake for validation, as well as for testing. For transfer learning, we used 10 real and 10 fake videos.
}
\label{tab:dataset_details}
\resizebox{1\columnwidth}{!}{%
\begin{tabular}{|l|r|r|r|r|} 
\hline
\multicolumn{1}{|c|}{ \textbf{Datasets} } & \multicolumn{1}{c|}{\begin{tabular}[c]{@{}c@{}}\textbf{Total}\\\textbf{ Videos} \end{tabular}} & \multicolumn{1}{c|}{\begin{tabular}[c]{@{}c@{}}\textbf{Base}\\\textbf{Training}\\\textbf{videos} \end{tabular}} & \multicolumn{1}{c|}{\begin{tabular}[c]{@{}c@{}}\textbf{Transfer}\\\textbf{Learning}\\\textbf{videos} \end{tabular}} & \multicolumn{1}{c|}{\begin{tabular}[c]{@{}c@{}}\textbf{Samples}\\\textbf{per}\\\textbf{video} \end{tabular}} \\ 
\hhline{=====}
 \textbf{Pristine (Real)}  & 1,000 & 750 & 10 & 16 \\ 
\hline
\textbf{DeepFake (DF)}  & 1,000 & 750 & 10 & 16 \\ 
\hline
\textbf{FaceSwap (FS)}  & 1,000 & 750 & 10 & 16 \\ 
\hline
\textbf{Face2Face (F2F)}  & 1,000 & 750 & 10 & 16 \\ 
\hline
\textbf{NeuralTextures (NT)}  & 1,000 & 750 & 10 & 16 \\ 
\hline
\textbf{DeepFakeDetection (DFD)}  & 3,363 & 250 & 10 & 16 \\ 
\hline
\end{tabular}
}
\end{table}
\subsubsection{ShallowNet.} Tariq\textit{ et al.}~\cite{Shahroz2} showed that ShallowNet~\cite{Shahroz1} achieves high accuracy in detecting computer-generated images. We developed ShallowNet using Python v3.6.8 using TensorFlow v1.14.0~\cite{tensorflow} and used the Keras v2.2.4~\cite{Keras}.

\subsubsection{FaceForensics++ (FF++).} R\"ossler\textit{ et al.}~\cite{FaceForensics++} used a modified version of the Xception Network to detect DeepFake, FaceSwap, Face2Face, and NeuralTextures. We are directly using results from FaceForensics++~\cite{FaceForensics++}, since they used the same dataset.

\subsubsection{DenseNet with Bidirectional RNN.} Sabir\textit{ et al.}~\cite{DFD2} used DenseNet with a bidirectional RNN to achieve high accuracy on DeepFake, FaceSwap, and Face2Face datasets. Similar to our CLRNet, this work also uses five consecutive frames for the training and testing of the model. We are directly using the results of Sabir\textit{ et al.}~\cite{DFD2}, since they used the same dataset. 

\subsubsection{Forensics Transfer (FT).} Cozzolino\textit{ et al.}~\cite{ForensicTransfer} developed a weakly supervised method for domain adaptation using an autoencoder based approach. They divide the latent space into real and fake parts to achieve higher detection accuracy on the Face2Face and FaceSwap datasets. For the implementation of ForensicTransfer autoencoder~\cite{ForensicTransfer}, we used PyTorch v1.1.0 ~\cite{PyTorch} on Python v3.6.8.

\begin{figure}[t!]
    \begin{subfigure}{0.48\columnwidth}
        \centering
        \includegraphics[clip, trim=1.1cm 1.3cm 0.3cm 0.5cm, width=1\columnwidth]{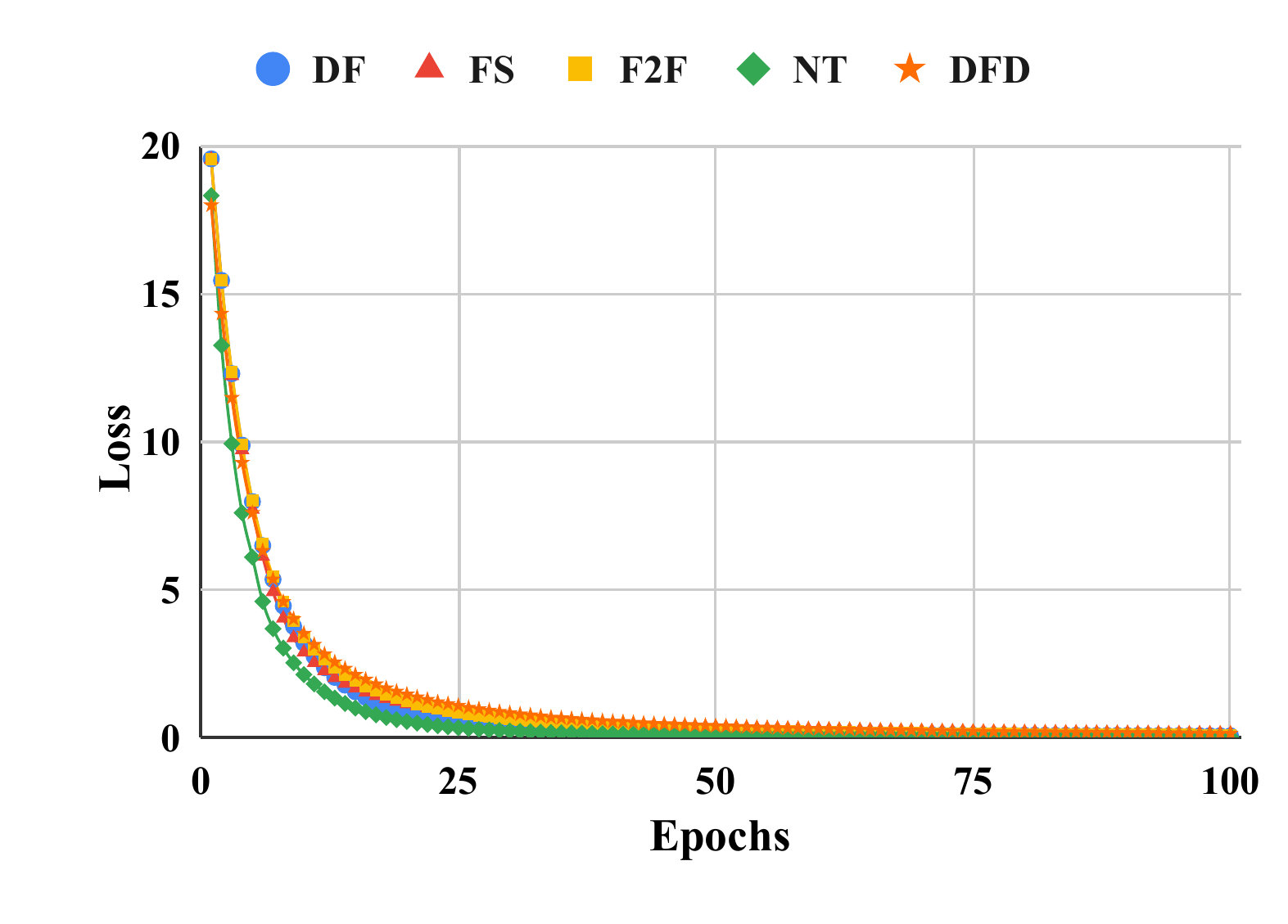}
        \caption{Training Loss}
    \end{subfigure}\quad
    \begin{subfigure}{0.48\columnwidth}
        \centering
        \includegraphics[clip, trim=1.1cm 1.3cm 0.3cm 0.5cm, width=1\columnwidth]{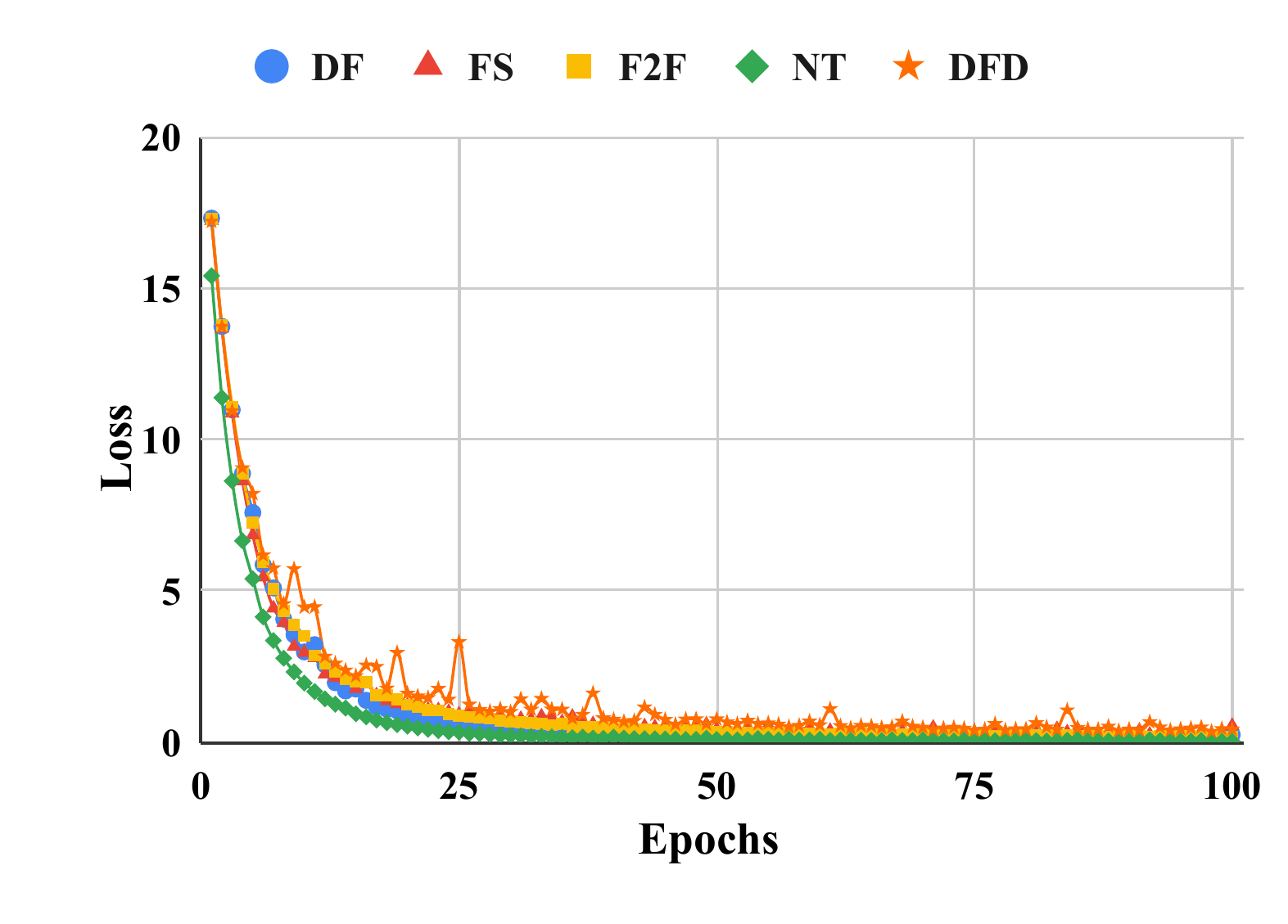}
        \caption{Validation Loss}
    \end{subfigure}
    \caption{\emph{Training vs Validation Loss of CLRNet}: Training and validation losses of CLRNet on the DeepFake (DF), FaceSwap(FS), Face2Face (F2F), NeuralTextures (NT), and DeepFakeDetection (DFD) datasets. The loss progression for both training and validation is descending very similarly, which shows that our model is learning accurately and is not overfitting to the training data. }
    \label{fig:Trainingloss}
\end{figure}
\section{Results}
\label{sec:res}
We have performed extensive experiments to evaluate and compare the performances of CLRNet and baseline methods. Only the most important experiments and their findings are discussed in this paper. 
The following sections will discuss the results from different experiments in detail.
\begin{figure*}[t]
    \begin{subfigure}[t]{0.32\linewidth}
        \centering
        \includegraphics[clip, trim=1.1cm 0.9cm 0.3cm 0.5cm,width=1\columnwidth]{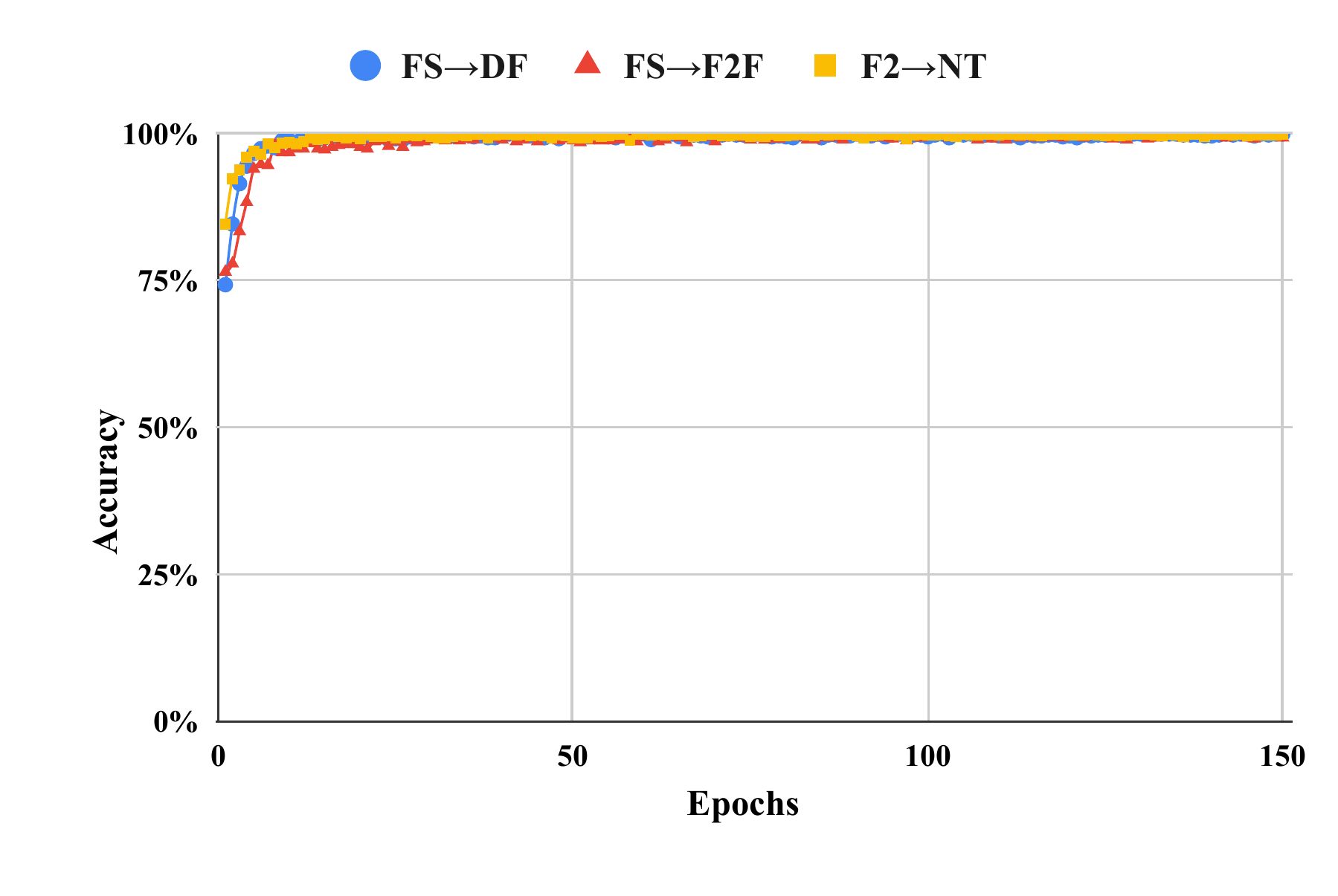}
        \caption{\emph{Training (Source $+$ Target)}: We used ten videos from source and ten from the target dataset. CLRNet shows high learning capability, even with such small amount of data.}
    \end{subfigure}\quad
    \begin{subfigure}[t]{0.32\linewidth}
        \centering
        \includegraphics[clip, trim=1.1cm 0.9cm 0.3cm 0.5cm,width=1\columnwidth]{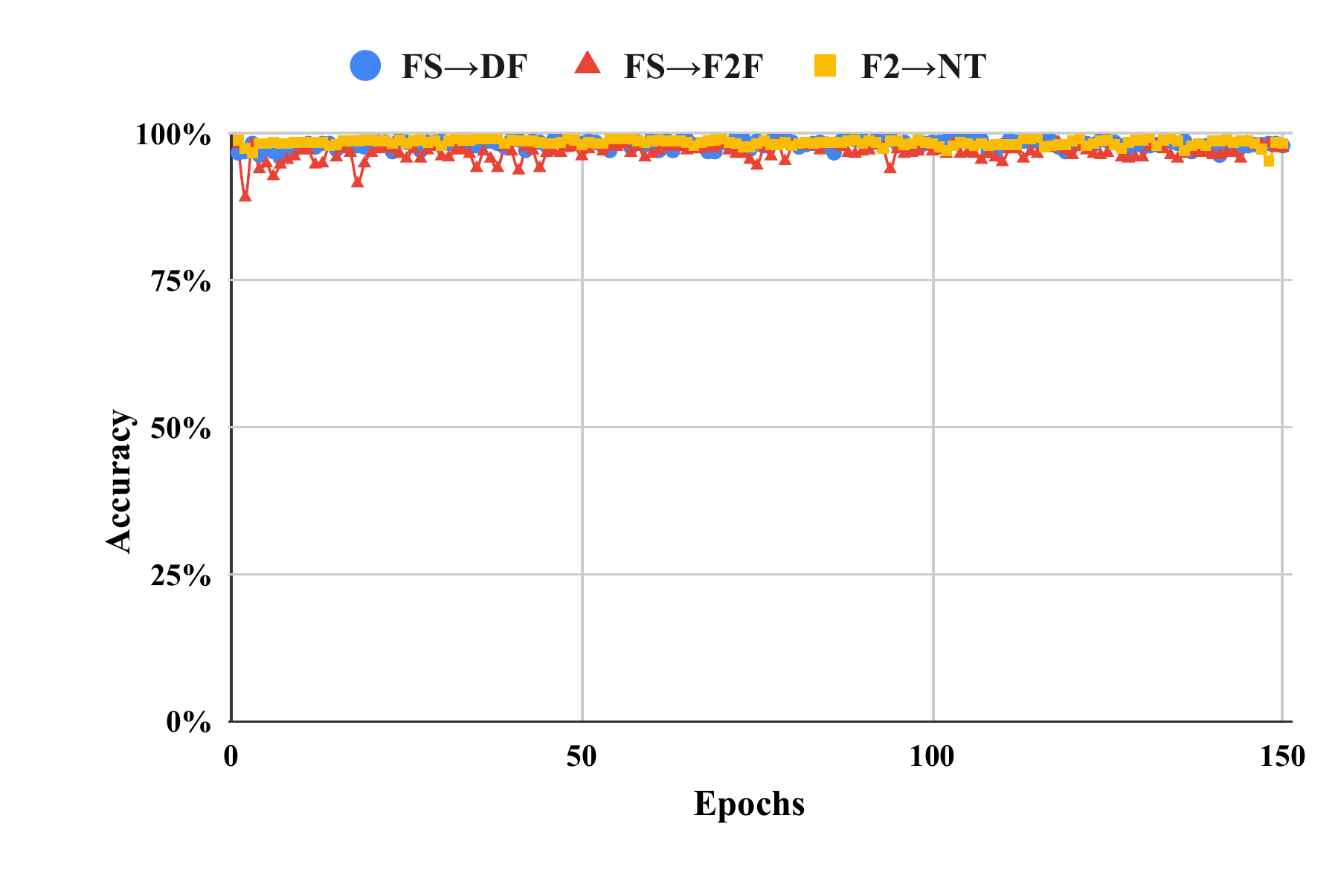}
        \caption{\emph{Validation (Source)}: We achieved high accuracy on the source dataset by providing ten videos from the source dataset during transfer learning.}
    \end{subfigure}\quad
    \begin{subfigure}[t]{0.32\linewidth}
        \centering
        \includegraphics[clip, trim=1.1cm 0.9cm 0.3cm 0.5cm,width=1\columnwidth]{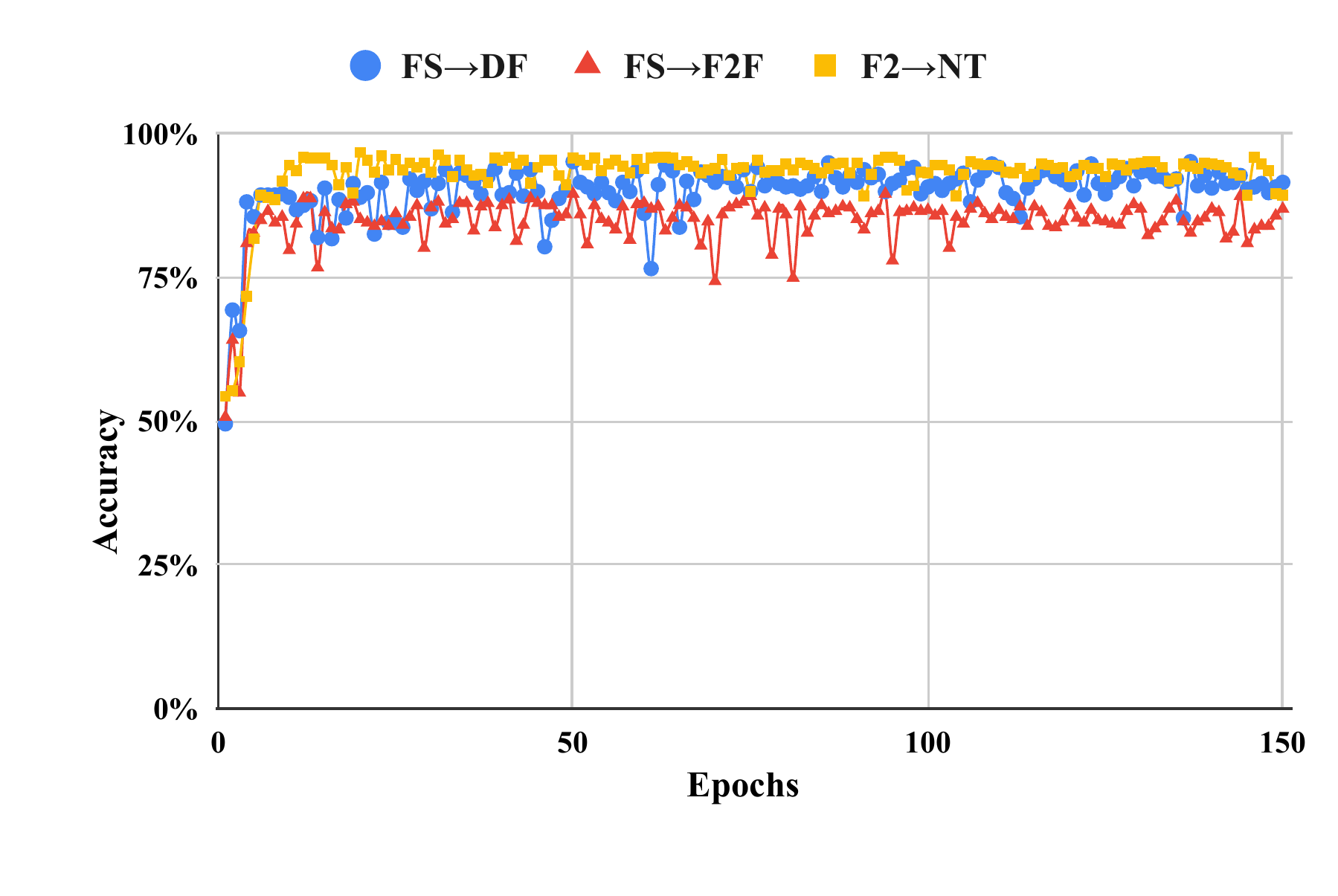}
        \caption{\emph{Validation (Target)}: We also achieved relatively high accuracy on the target dataset by using only ten videos for the target dataset with CLRNet.}
    \end{subfigure}
    \caption{\emph{Transfer Learning Accuracy of CLRNet}: Comparison of (a) transfer learning (TL) training accuracy with (b) the validation accuracy of the source and (c) the validation accuracy of the target. The arrow `$\to$' represents the TL from a source dataset to a target dataset.}
    \label{fig:TLloss}
\end{figure*}
\subsection{Learning Capability of CLRNet}
Figure~\ref{fig:Trainingloss} presents the training and validation losses of CLRNet on different datasets (DF, FS, F2F, NT, and DFD). We can observe that the training and validation lossess gradually descend in a similar fashion, which shows that our CLRNet model is not overfitting to the training dataset and is learning distinguishable features between real and fake images from the training set. As shown in Fig.~\ref{fig:Trainingloss}, there are slight fluctuations in the validation loss of DFD, but it is not the case for other datasets. We believe that this is due to the smaller size of the dataset used for DFD training (250 videos) as compared to other datasets (750 videos), as shown in Table~\ref{tab:dataset_details}. Another cause is the dynamic environment in DFD videos, making it harder for the model to learn. However, our CLRNet model was able to learn from the DFD dataset, even with a smaller data size, and achieved a 96.00\% F1-score, as shown in Table~\ref{tab:training}. 
\subsection{Performance on Base Dataset}
We trained our CLRNet and other baseline models on a base training dataset (DF or FS), as shown in Table~\ref{tab:training}. Then we compared the performance of the trained model on a test set from the base dataset and also evaluated the zero-shot performance from the test set of other datasets. The baseline methods (ShallowNet, Xception, and FT) and CLRNet are trained and evaluated on the set of images from the datasets. However, the results of FF++~\cite{FaceForensics++} and Sabir\textit{ et al.}~\cite{DFD2} are directly taken from their work. Table~\ref{tab:training} reports all the results. As we can observe from Table~\ref{tab:training}, when trained on the DeepFake base dataset, FT (99.35\%) outperforms CLRNet (99.02\%) by a slight margin. The performances of our vanilla implementation of Xception (86.00\%) and ShallowNet (56.65\%) do not bear good results. However, the modified versions of Xception from FF++ (96.36\%) and DenseNet of Sabir\textit{ et al.} (96.90\%) perform better than ShallowNet (56.65\%) and the vanilla Xception (86.00\%). The zero-shot performance is relatively low (below 85\%) for all methods: FT performed the best on F2F (74.12\%) and NT (83.54\%), and CLRNet performed the best on DFD (60.38\%).
When trained with the FaceSwap (FS) base dataset, CLRNet (FS: 98.05\%) is the best performer as compared to the best baseline method FT (FS: 96.17\%), as shown in Table~\ref{tab:training}. Similar to the DF base dataset, Xception (FS: 85.37\%), FF++ (FS: 90.29\%), and Sabir el al. (FS: 94.35\%) showed decent performance, and ShallowNet showed the worst performance (FS: 52.75\%). We also evaluated our CLRNet method on NeuralTextures (NT) and DeepFakeDetection (DFD) base datasets. As shown in Table~\ref{tab:training}, CLRNet achieves a very high F1-score for both datasets (NT: 99.50\%, DFD: 96.00\%). From Table~\ref{tab:training}, we can observe that there is no best method in terms of the zero-shot performance. Therefore, we explore transfer learning in the next experiments.
\vspace{-1.7mm}
\subsection{Transfer Learning with Source and Target}
In our preliminary experiments, we observed that when the model is trained with only the target dataset for transfer learning, its performance on the source dataset drops (as low as 50\%), which is not ideal for building a generic deepfake detector. Therefore, we combine a small volume of the source (10 videos) and target (10 videos) datasets for transfer learning, so that the model can remember the features of the source dataset as well. Figure~\ref{fig:TLloss} shows the training accuracy (source $+$ target), validation accuracy on the source dataset (FS), and validation accuracy on the target dataset (DF or F2F or NT). We can observe from Fig.~\ref{fig:TLloss} that the validation accuracy of the source dataset (FS) slightly drops (1\textasciitilde2\%), and the validation accuracy on the target dataset (DF or F2F or NT) increases over time (reaches up to 90\%). This technique allows CLRNet to achieve high performance on transfer learning for both the source and target datasets. In the next section, we will discuss the results of transfer learning. 
\subsection{Transfer Learning Performance}
For transfer learning, we used only CLRNet and the best performing baseline model on the base dataset, namely FT. Cozzolino\textit{ et al.}~\cite{ForensicTransfer} performed only the Face2Face (source) to FaceSwap (target) experiment for the FF++ dataset~\cite{FaceForensics++}. We performed extensive experiments with FT and compared it with our CLRNet model. We performed three different types of experiments, each consisting of multiple sub-experiments, for the transfer learning (TL) task. The following sections provide the details of these experiments and the performance evaluation of CLRNet and FT.

\subsubsection{Exp 1. Single--source to Single--target.}
In this experiment, we used one source dataset (either FS, F2F, NT, DFD or DF) and one target dataset (either FS, F2F, NT, DFD or DF) for transfer learning and compared the performance of FT and CLRNet. Due to space limitations, we partially report this experiment in Table~\ref{tab:TL}.
In experiment 1, we set DeepFake (DF) as the source and FaceSwap (FS) as the target. The CLRNet accuracy increases from 50.00\% to 83.08\% for the target (FS) and drops from 99.02\% to 90.95\% for source (DF), as shown in Table~\ref{tab:training} and~\ref{tab:TL}. In contrast, FT accuracy drops from 50.00\% (FS) and 99.35\% (DF) to 47.72\% (FS) and 62.59\% (DF), respectively, as shown in Table~\ref{tab:training} and ~\ref{tab:TL}. Similarly, when Face2Face (F2F) is the target, the CLRNet accuracy increases from 53.73\% to 88.35\% for the target, whereas the FT accuracy increases only by 5\% from 74.12\% to 79.31\%. This low performance demonstrates the limitation of FT regarding generalization.
In experiment 2, we set FaceSwap (FS) as the source, and the performance for the target datasets (DF and F2F) are even better than in experiment 1, in favor of CLRNet. When DeepFake (DF) is the target, CLRNet (FS: 96.33\%, DF: 92.47\%) outperforms FT (FS: 44.93\%, DF:76.94\%) by 50\% for FS and 16\% for DF, respectively. Also, when Face2Face (F2F) is the target, CLRNet (FS: 93.93\%, F2F: 87.48\%) outperforms FT (FS: 55.74\%, F2F: 53.73\%), as shown in Table~\ref{tab:TL}. We also experimented with CLRNet by setting the target as NeuralTextures (NT) or DeepFakeDetection (DFD). After transfer learning, the performance on NT increases from 98.05\% to 98.12\%  for the source (FS) and from 53.33\% to 95.50\% for the target (NT), as shown in Table~\ref{tab:training} and ~\ref{tab:TL}. Finally, for DFD, we observed a similar trend and achieved an increase from 49.13\% to 88.88\%.

Based on our experiment, we observed that the accuracy of CLRNet always increases for the target as compared to its zero-shot performance with a slight decrease or sometimes an increase in the source accuracy, as shown in Table~\ref{tab:training} and ~\ref{tab:TL}. On the other hand, the performance of FT is very unstable and generally worse than the single dataset performance, which shows significant limitation of FT; CLRNet overcomes this limitation and achieves better performance.
\begin{table}
\centering
\caption{\emph{The base dataset and zero-shot performance}: This table shows the performance comparison of our CLRNet with five different baseline models. First, we trained the models on a base dataset. Afterward, we tested all models on the base dataset (highlighted in gray) and on other datasets as well (zero-shot). Our CLRNet has the highest performance for all base datasets except one. However, for zero-shot performance, CLRNet performed the best on FS and FT as well as on DF and F2F. `$\ddagger$' represents our implementation of the method.}
\label{tab:training}
\resizebox{\linewidth}{!}{%
\begin{tabular}{|l|c|l|l|l|l|l|} 
\hline
\multirow{2}{*}{ \textbf{Method} } & \multirow{2}{*}{\begin{tabular}[c]{@{}c@{}}\textbf{Base}\\\textbf{ Dataset} \end{tabular}} & \multicolumn{1}{c|}{\textbf{DF} } & \multicolumn{1}{c|}{\textbf{FS} } & \multicolumn{1}{c|}{\textbf{F2F} } & \multicolumn{1}{c|}{\textbf{NT} } & \textbf{DFD}  \\
 &  & \multicolumn{1}{c|}{\textbf{(\%)}} & \multicolumn{1}{c|}{\textbf{(\%)}} & \multicolumn{1}{c|}{\textbf{(\%)}} & \multicolumn{1}{c|}{\textbf{(\%)}} & \multicolumn{1}{c|}{\textbf{(\%)}} \\ 
\hhline{=======}
ShallowNet$^\ddagger$ & \multirow{6}{*}{DF} & {\cellcolor[rgb]{0.902,0.902,0.902}}56.65 & \textbf{50.32 } & 54.15 & 35.13 & 41.23 \\ 
\hhline{|-~-----|}
Xception &  & {\cellcolor[rgb]{0.902,0.902,0.902}}86.00 & 49.78 & 57.26 & 61.49 & \textbf{68.10} \\ 
\hhline{|-~-----|}
FF++~\cite{FaceForensics++} &  & {\cellcolor[rgb]{0.902,0.902,0.902}}96.36 & \multicolumn{1}{c|}{-} & \multicolumn{1}{c|}{-} & \multicolumn{1}{c|}{-} & \multicolumn{1}{c|}{-} \\ 
\hhline{|-~-----|}
Sabir\textit{ et al.}~\cite{DFD2} &  & {\cellcolor[rgb]{0.902,0.902,0.902}}96.90 & \multicolumn{1}{c|}{-} & \multicolumn{1}{c|}{-} & \multicolumn{1}{c|}{-} & \multicolumn{1}{c|}{-} \\ 
\hhline{|-~-----|}
FT$^\ddagger$ &  & {\cellcolor[rgb]{0.902,0.902,0.902}}\textbf{99.35}  & 50.00 & \textbf{74.12}  & \textbf{83.54}  & 48.45 \\ 
\hhline{|-~-----|}
\textbf{CLRNet (Ours)}  &  & {\cellcolor[rgb]{0.902,0.902,0.902}}99.02 & 50.00 & 53.73 & 69.75 & 60.38 \\ 
\hhline{=======}
ShallowNet$^\ddagger$ & \multirow{6}{*}{FS} & 47.09 & {\cellcolor[rgb]{0.902,0.902,0.902}}52.75 & 47.50 & 46.96 & \textbf{49.58} \\ 
\hhline{|-~-----|}
Xception &  & \textbf{49.49} & {\cellcolor[rgb]{0.902,0.902,0.902}}85.37 & \textbf{56.61} & 52.05 & 48.08 \\ 
\hhline{|-~-----|}
FF++~\cite{FaceForensics++} &  & \multicolumn{1}{c|}{-} & {\cellcolor[rgb]{0.902,0.902,0.902}}90.29 & \multicolumn{1}{c|}{-} & \multicolumn{1}{c|}{-} & \multicolumn{1}{c|}{-} \\ 
\hhline{|-~-----|}
Sabir\textit{ et al.}~\cite{DFD2} &  & \multicolumn{1}{c|}{-} & {\cellcolor[rgb]{0.902,0.902,0.902}}94.35 & \multicolumn{1}{c|}{-} & \multicolumn{1}{c|}{-} & \multicolumn{1}{c|}{-} \\ 
\hhline{|-~-----|}
FT$^\ddagger$ &  & 48.86 & {\cellcolor[rgb]{0.902,0.902,0.902}}96.17 & 54.39 & 47.45 & 36.66 \\ 
\hhline{|-~-----|}
\textbf{CLRNet (Ours)}  &  & 48.53 & {\cellcolor[rgb]{0.902,0.902,0.902}}\textbf{98.05}  & 50.15 & \textbf{53.33}  & 49.13 \\ 
\hhline{=======}

FF++~\cite{FaceForensics++} & \multirow{2}{*}{NT} & \multicolumn{1}{c|}{-} & \multicolumn{1}{c|}{-} & \multicolumn{1}{c|}{-} & \multicolumn{1}{S[table-format=2.2]|}{{\cellcolor[rgb]{0.902,0.902,0.902}}80.67} & \multicolumn{1}{c|}{-} \\ 
\hhline{|-~-----|}
\textbf{CLRNet (Ours)} &  & \textbf{50.12} & \textbf{49.93} & \textbf{49.80} & {\cellcolor[rgb]{0.902,0.902,0.902}}\textbf{99.50} & \textbf{50.00} \\ 
\hhline{=======}
\textbf{CLRNet (Ours)}  & DFD & \textbf{65.08}  & \textbf{51.92}  & \textbf{60.32}  & \textbf{63.10}  & {\cellcolor[rgb]{0.902,0.902,0.902}}\textbf{96.00}  \\
\hline
\end{tabular}%
}
\end{table}
\subsubsection{Exp 2. Multi--source to Single--target.}
In this experiment, we first trained the model with two different source datasets and then performed transfer learning to a new target dataset. In Table~\ref{tab:TL}, we report the best performing model of this experiment for both FT and CLRNet, namely FS+DF (source) and F2F (target). The accuracy of CLRNet (FS: 94.37\%, DF: 92.15\%) on the source is significantly higher compared to that of FT (FS: 45.17\%, DF: 64.08\%). Similarly, CLRNet(F2F: 86.22\%) outperforms FT (F2F: 55.98\%) on the target dataset, as shown in Table~\ref{tab:TL}. CLRNet also exceeds the performance of FT in terms of zero-shot learning on NT (CLRNet: 79.75\%, FT: 62.09\%) and DFD (CLRNet: 63.12\%, FT: 51.38\%). This experiment also validates the superiority of CLRNet over FT at deepfake detection generalization.
\subsubsection{Exp. 3. Single--source to Multi--target.}
In this experiment, we test the generalizability of our CLRNet model, that is, its performance to detect multiple deepfake types at once with only a small amount of data. For this experiment, we first trained our CLRNet model on a source dataset (FS) and then performed three types of experiments.
In experiment 1, we set the target as DF+F2F and evaluated the accuracy of the model after transfer learning. From Table~\ref{tab:TL}, we can observe the performance for the source (FS: 94.55\%) and the target (DF: 91.77\%, F2F:87.75\%). In experiment 2, we set the target as DF+F2F+NT. From Table~\ref{tab:TL}, we can see the accuracy for the source (FS: 94.35\%) and the target (DF: 90.67\%, F2F:85.78\%, NT: 91.47\%). In this experiment, we set the target as DF+F2F+NT+DFD. In Table~\ref{tab:TL}, we can see the accuracy for the source (FS: 93.70\%) and the target (DF: 91.23\%, F2F:87.50\%, NT: 91.30\%, DFD: 87.13\%). Based on these experiments, we concluded: 1) that by increasing the number of target datasets, the source dataset accuracy slightly decreases (1\textasciitilde2\%), and that 2) CLRNet generalizes well on different deepfake datasets, since its performance is better than the zero-shot performance on the same dataset in all of our experiments.

\begin{table}
\centering
\caption{\emph{Transfer Learning Performance}: This table shows the performance comparison of our CLRNet model and the best baseline method from the previous experiment, namely FT. In this experiment, we first trained the model on a source dataset and then used a small target dataset (highlighted in gray) consisting of 10 videos to perform transfer learning. Afterward, we performed testing on all datasets. We performed ten experiments belonging to 3 different types: 1) Single-source to single-target, 2) Multi-source to single target, and 3) Single-source to Multi-target. Our CLRNet model performed the best in all scenarios. `$\ddagger$' represents our implementation of the method.}
\label{tab:TL}
\resizebox{\linewidth}{!}{%
\begin{tabular}{|l|c|c|l|l|l|l|l|} 
\hline
\multirow{2}{*}{ \textbf{Method} } & \multirow{2}{*}{\textbf{Source} } & \multirow{2}{*}{\textbf{Target} } & \multicolumn{1}{c|}{\textbf{DF} } & \multicolumn{1}{c|}{\textbf{FS} } & \multicolumn{1}{c|}{\textbf{F2F} } & \multicolumn{1}{c|}{\textbf{NT} } & \textbf{DFD}  \\
 &  &  & \multicolumn{1}{c|}{(\%)} & \multicolumn{1}{c|}{(\%)} & \multicolumn{1}{c|}{(\%)} & \multicolumn{1}{c|}{(\%)} & \multicolumn{1}{c|}{(\%)} \\ 
\hhline{========}
\multicolumn{8}{|c|}{\textbf{Single--source to Single--target}} \\ 
\hline
FT$^\ddagger$ & \multirow{4}{*}{DF} & \multirow{2}{*}{FS} & 62.59 & {\cellcolor[rgb]{0.902,0.902,0.902}}47.72 & \textbf{61.34}  & \textbf{68.51}  & 51.38 \\ 
\hhline{|-~~-----|}
\textbf{CLRNet}  &  &  & \textbf{90.95}  & {\cellcolor[rgb]{0.902,0.902,0.902}}\textbf{83.08}  & 48.68 & 65.00 & \textbf{53.87}  \\ 
\hhline{|=~======|}
FT$^\ddagger$ &  & \multirow{2}{*}{F2F} & 83.70 & \textbf{54.92}  & {\cellcolor[rgb]{0.902,0.902,0.902}}79.31 & \textbf{82.24}  & 54.92 \\ 
\hhline{|-~~-----|}
\textbf{CLRNet}  &  &  & \textbf{97.18}  & 49.28 & {\cellcolor[rgb]{0.902,0.902,0.902}}\textbf{88.35} & 78.05 & \textbf{68.13}  \\ 
\hline
FT$^\ddagger$ & \multirow{8}{*}{FS} & \multirow{2}{*}{DF} & {\cellcolor[rgb]{0.902,0.902,0.902}}76.94 & 44.93 & \textbf{66.42}  & 70.53 & 49.78 \\ 
\hhline{|-~~-----|}
\textbf{CLRNet}  &  &  & {\cellcolor[rgb]{0.902,0.902,0.902}}\textbf{92.47}  & \textbf{96.33}  & 65.58 & \textbf{75.80}  & \textbf{59.13}  \\ 
\hhline{|=~======|}
FT$^\ddagger$ &  & \multirow{2}{*}{F2F} & 56.06 & 55.74 & {\cellcolor[rgb]{0.902,0.902,0.902}}53.73 & 55.38 & 47.63 \\ 
\hhline{|-~~-----|}
\textbf{CLRNet}  &  &  & \textbf{79.87}  & \textbf{93.93}  & {\cellcolor[rgb]{0.902,0.902,0.902}}\textbf{87.48}  & \textbf{78.00}  & \textbf{52.50}  \\ 
\hhline{|=~======|}
FT$^\ddagger$ &  & \multirow{2}{*}{NT} & \multicolumn{1}{c|}{-} & \multicolumn{1}{c|}{-} & \multicolumn{1}{c|}{-} & \multicolumn{1}{c|}{{\cellcolor[rgb]{0.902,0.902,0.902}}-} & \multicolumn{1}{c|}{-} \\ 
\hhline{|-~~-----|}
\textbf{CLRNet}  &  &  & 51.82 & 98.12 & 50.90 & {\cellcolor[rgb]{0.902,0.902,0.902}}95.50 & 49.13 \\ 
\hhline{|=~======|}
FT$^\ddagger$ &  & \multirow{2}{*}{DFD} & \multicolumn{1}{c|}{-} & \multicolumn{1}{c|}{-} & \multicolumn{1}{c|}{-} & \multicolumn{1}{c|}{-} & \multicolumn{1}{c|}{{\cellcolor[rgb]{0.902,0.902,0.902}}-} \\ 
\hhline{|-~~-----|}
\textbf{CLRNet}  &  &  & 70.97 & 96.15 & 51.85 & 77.02 & {\cellcolor[rgb]{0.902,0.902,0.902}}88.88 \\ 
\hhline{========}
\multicolumn{8}{|c|}{\textbf{Multi--source to Single--target }} \\ 
\hline
FT$^\ddagger$ & \multirow{2}{*}{FS$+$DF } & \multirow{2}{*}{F2F} & 64.08 & 45.17 & {\cellcolor[rgb]{0.902,0.902,0.902}}55.98 & 62.09 & 51.38 \\ 
\hhline{|-~~-----|}
\textbf{CLRNet}  &  &  & \textbf{92.15}  & \textbf{94.37}  & {\cellcolor[rgb]{0.902,0.902,0.902}}\textbf{86.22}  & \textbf{79.75}  & \textbf{63.12}  \\ 
\hhline{========}
\multicolumn{8}{|c|}{\textbf{Single--source to Multi--target }} \\ 
\hline
\multicolumn{1}{|c|}{\multirow{5}{*}{\begin{tabular}[c]{@{}c@{}}\textbf{CLRNet}\\\textbf{(best)} \end{tabular}}} & \multirow{5}{*}{FS} & DF$+$F2F & {\cellcolor[rgb]{0.902,0.902,0.902}}\textbf{91.77 } & \textbf{94.55 } & {\cellcolor[rgb]{0.902,0.902,0.902}}\textbf{87.75 } & 85.12 & 69.13 \\ 
\hhline{|~~------|}
\multicolumn{1}{|c|}{} &  & DF$+$F2F & \multicolumn{1}{c|}{{\cellcolor[rgb]{0.902,0.902,0.902}}} & \multicolumn{1}{c|}{\multirow{2}{*}{94.35}} & \multicolumn{1}{c|}{{\cellcolor[rgb]{0.902,0.902,0.902}}} & \multicolumn{1}{c|}{{\cellcolor[rgb]{0.902,0.902,0.902}}} & \multicolumn{1}{c|}{\multirow{2}{*}{69.50}} \\
\multicolumn{1}{|c|}{} &  & $+$NT & \multicolumn{1}{c|}{\multirow{-2}{*}{{\cellcolor[rgb]{0.902,0.902,0.902}}90.67}} & \multicolumn{1}{c|}{} & \multicolumn{1}{c|}{\multirow{-2}{*}{{\cellcolor[rgb]{0.902,0.902,0.902}}85.78}} & \multicolumn{1}{c|}{\multirow{-2}{*}{{\cellcolor[rgb]{0.902,0.902,0.902}}\textbf{91.47 }}} & \multicolumn{1}{c|}{} \\ 
\hhline{|~~------|}
\multicolumn{1}{|c|}{} &  & DF$+$F2F & \multicolumn{1}{c|}{{\cellcolor[rgb]{0.902,0.902,0.902}}} & \multicolumn{1}{c|}{\multirow{2}{*}{93.70}} & \multicolumn{1}{c|}{{\cellcolor[rgb]{0.902,0.902,0.902}}} & \multicolumn{1}{c|}{{\cellcolor[rgb]{0.902,0.902,0.902}}} & \multicolumn{1}{c|}{{\cellcolor[rgb]{0.902,0.902,0.902}}} \\
\multicolumn{1}{|c|}{} &  & $+$NT$+$DFD & \multicolumn{1}{c|}{\multirow{-2}{*}{{\cellcolor[rgb]{0.902,0.902,0.902}}91.23}} & \multicolumn{1}{c|}{} & \multicolumn{1}{c|}{\multirow{-2}{*}{{\cellcolor[rgb]{0.902,0.902,0.902}}87.50}} & \multicolumn{1}{c|}{\multirow{-2}{*}{{\cellcolor[rgb]{0.902,0.902,0.902}}91.30}} & \multicolumn{1}{c|}{\multirow{-2}{*}{{\cellcolor[rgb]{0.902,0.902,0.902}}\textbf{87.13 }}} \\
\hline
\end{tabular}%
}
\end{table}

\section{Conclusion}
\label{sec:conclusion}
We introduced CLRNet, which is successfully applied to detect a variety of deepfake videos. Instead of using a single frame, our model uses a sequence of consecutive frames from the video as an input, which helps our CLRNet model to capture and incorporate the temporal information and detect artifacts present between consecutive frames. Through more than 5 different experiments, we have shown the superiority of our CLRNet model over the previous baselines and state-of-the-art methods. To conclude, we addressed the shortcomings of the previous state-of-the-art methods by proposing a more generalizable model with a better detection performance. 
From this work, we hope that our research is a stepping stone toward developing more generalized deepfake detectors and future work will continue to challenge and improve existing deepfake detection methods to more generalizable and universal approaches.

\bibliographystyle{ACM-Reference-Format}
\interlinepenalty=10000
\bibliography{references.bib}
\clearpage
\appendix

\section{Appendix on Reproducibility}
\label{sec:repo}

This section describes all the necessary information required to reproduce the results from the experiments for CLRNet and all baselines methods. Here, we tried our best to cover everything to aid in reproducibility.

\subsection{Hardware and software Configuration}
We used Intel(R) Xeon(R) Silver 4114 CPU @ 2.20GHz with 256.0GB RAM and NVIDIA GeForce Titan RTX. The following package versions were used on Python v3.7.5: TensorFlow-gpu: v1.13.1; Keras-applications: v1.0.8; Keras-preprocessing: v1.1.0; cudatoolkit: v10.0.130; numpy: v1.17.4; Pandas: v0.25.3; Opencv: 3.4.2; scikit-learn: v0.21.3; PyTorch: v1.1.0.

\subsection{Deepfake Video Dataset}
We downloaded the dataset (DeepFake, Face2Face, FaceSwap, NeuralTextures, and DeepFakeDetection) from the FaceForensics++ GitHub repository\footnote{\href{https://github.com/ondyari/FaceForensics}{https://github.com/ondyari/FaceForensics}} for the training and evaluation of the baseline and our CLRNet model. 

\subsection{Implementation of CLRNet}
We implemented our CLRNet model using TensorFlow and Keras on Python. All the layers used in the model are available in the Keras library. A detailed view of CLRNet architecture is provided in Table~\ref{tab:CLRNet_Details}. We used the Keras convention to name the layers. The input shape for our model is (Videos, Samples, Rows, Columns, Channels). The ImageDataGenerator in Keras can load data of the form (Samples, Rows, Columns, Channels). Therefore, we could not use the ImageDataGenerator provided by the Keras library. So we built our own VideoDataGenerator with very similar functionality as that of ImageDataGenerator. However, as we planned to provide the input in a set of 5 consecutive frames, we had to implement some extra features as well. All of this is present in our source code, which we plan to release along with this paper.

\subsection{Baseline Model Implementation}
We compared our CLRNet model with five baseline models (Xception, ShallowNet, FF++, DenseNet with Bidirectional RNN, and FT). Below, we explain how these baselines are implemented.

\subsubsection{Xception.}
We used the Keras implementation of Xception, which is pre-trained on the ImageNet dataset.

\subsubsection{ShallowNet.} We implemented ShallowNetV3 based on the specification provided by Tariq\textit{ et al.}~\cite{Shahroz1} The architecture of ShallowNetV3 is shown in Table~\ref{tab:ShallowNet}. ShallowNetV3 consists of 4 blocks with 33 layers. One row in Table~\ref{tab:ShallowNet} correspond to one block. We implemented this architecture using Keras and TensorFlow as backend.

\subsubsection{ForensicTransfer (FT).} We implemented ShallowNetV3 based on the specification provided by Cozzolino\textit{ et al.}~\cite{ForensicTransfer} The architecture of the ForensicsTransfer (FT) model is shown in Table~\ref{tab:FT_Arc}. We implemented the FT model using PyTorch v.1.2.0. We tried our best to mimic the original performance of the baseline.

\subsection{Data preparation for CLRNet}
For dataset preparation, we randomly extracted 5 consecutive frames from each video and used them as one sample. We extracted 16 samples from each video. The video whose frames are used in training has not been used for validation or testing. We cropped the faces from the frames using MTCNN~\cite{mtcnn}. We used our VideoDataGenerator to load these samples and feed them to CLRNet.

\subsection{Data Augumentation}
In Keras, the data augmentation feature can be used with the ImageDataGenerator. As we have implemented the VideoDataGenerator, we had to implement our own DataAugumentor as well. Therefore we used the Keras ImageDataGenerator Interface and extended it to VideoDataGenerator. The transformation setting we used for data augmentation are as follows: rotation\_range=30; brightness\_range = [0.7,1.0]; channel\_shift\_range=50.0; zoom\_range=0.2; horizontal\_flip=True; fill\_mode=`nearest'.

\subsection{Training of Models}
We used the Adam optimizer with the following setting: lr = 0.00005; beta\_1 = 0.9; beta\_2 = 0.999; epsilon = None; decay = 0.0; amsgrad = False. For the loss function, we used the Binary Cross-entropy. For every experiment, we trained our CLRNet and baseline models for 100 epochs. The best epoch based on the validation loss is used for evaluation with test datasets. We kept the same number of real and fake samples in training, validation, and test datasets. For the training of baseline methods, we used the same training and testing methods as specified in their original paper.

\subsection{Transfer Learning}
We freeze the first 120 layers of our CLRNet model during transfer learning. There is a small problem in Keras regarding the batch normalization layer when it comes to transfer learning. We solved this problem by using the solution provided by Vasilis Vryniotis\footnote{\href{http://blog.datumbox.com/the-batch-normalization-layer-of-keras-is-broken/}{http://blog.datumbox.com/the-batch-normalization-layer-of-keras-is-broken/}} and Oleg Gusev\footnote{\href{https://github.com/keras-team/keras/pull/9965\#issuecomment-549064001}{https://github.com/keras-team/keras/pull/9965\#issuecomment-549064001}}. For transfer learning, we used 150 epochs for each experiment, while keeping all other conditions the same. The number of datasets used to transfer a specific target is ten videos, and if there are two targets, ten videos from each target are used.

\subsection{Source Code}
We plan to release the source code for the implementations of CLRNet, our experiments, and ShallowNet and ForensicsTransfer (FT). If our paper gets accepted, we will clean up the code and share it our GitHub. The source code also contains Juypter Notebooks of all the experiments along with their results. 


\begin{table*}
\caption{ShallowNet Architecture. We used the following ShallowNetV3 architecture to develop this baseline model. Each row represents a block in the architecture. An L2 kernel regularizer of 0.0001 is used in each Conv2D layer.}
\label{tab:ShallowNet}
\centering
\begin{tabular}{|c|}
\hline
\textbf{ShallowNetV3} \\ \hline
Conv2D -- ReLU -- Dropout -- Conv2D -- ReLU -- Dropout -- Conv2D -- ReLU -- MaxPooling  -- Dropout \\ \hline
Conv2D -- ReLU -- Dropout -- Conv2D -- ReLU -- Dropout -- Conv2D -- ReLU -- MaxPooling -- Dropout  \\ \hline
Conv2D -- ReLU -- Dropout -- Conv2D -- ReLU -- Dropout \\ \hline
Flatten -- Dense -- ReLU -- BatchNormalization -- Dropout -- Dense -- Sigmoid  \\ \hline
\end{tabular}%
\end{table*}

\begin{table*}
\caption{ForensicsTransfer (FT) Architecture. We used the following architecture to develop the FT baseline model. Each Conv2d inside the Encoder part has a stride of 2, whereas each Conv2d in the Decoder part has a stride of 1. We used upsampling before each Conv2d layer inside the Decoder. }
\label{tab:FT_Arc}
\begin{tabular}{|l|c|}
\hline
\multicolumn{2}{|c|}{\textbf{ForensicsTransfer (FT)}} \\ \hline
\textbf{Encoder} & \begin{tabular}[c]{@{}c@{}}Conv2d -- ReLU -- Conv2d -- BatchNorm2d -- ReLU -- Conv2d -- BatchNorm2d -- ReLU -- Conv2d -- BatchNorm2d \\ -- ReLU -- Conv2d -- BatchNorm2d -- LReLU\end{tabular} \\ \hline
\textbf{Decoder} & \begin{tabular}[c]{@{}c@{}}Upsample -- Conv2d -- BatchNorm2d -- ReLU -- Upsample -- Conv2d -- BatchNorm2d -- ReLU -- Upsample -- Conv2d\\  -- BatchNorm2d -- ReLU -- Upsample  -- Conv2d -- BatchNorm2d -- ReLU -- ConvTranspose2d -- Tanh\end{tabular} \\ \hline
\end{tabular}%
\end{table*}

\begin{table*}
\centering
\caption{This is the detailed architecture of the CLRNet model that we used in this paper. All of these layers are available in the Keras library. The input size of the image is 240x240.}
\label{tab:CLRNet_Details}
\resizebox{0.70\linewidth}{!}{%
\begin{tabular}{|c|} 
\hline
 \textbf{Convolutional LSTM based Residual Network (CLRNet)}  \\ 
\hline
\begin{tabular}[c]{@{}c@{}}ConvLSTM2D -- BatchNorm -- ReLU -- MaxPooling3D\\ \end{tabular} \\ 
\hline
\begin{tabular}[c]{@{}c@{}}ConvLSTM2D -- Dropout~-- BatchNorm -- ReLU -- ConvLSTM2D -- Dropout -- BatchNorm~\\-- ConvLSTM2D~-- ReLU -- BatchNorm -- ConvLSTM2D~-- \textbf{Add}\\ \end{tabular} \\ 
\hline
\begin{tabular}[c]{@{}c@{}}BatchNorm -- ReLU -- ConvLSTM2D -- Dropout -- BatchNorm -- ReLU -- ConvLSTM2D -- Dropout \\-- BatchNorm~-- ReLU -- ConvLSTM2D -- \textbf{Add}\\ \end{tabular} \\ 
\hline
\begin{tabular}[c]{@{}c@{}}BatchNorm -- ReLU -- ConvLSTM2D -- Dropout -- BatchNorm -- ReLU -- ConvLSTM2D -- Dropout \\-- BatchNorm -- ReLU -- ConvLSTM2D -- \textbf{Add}\\ \end{tabular} \\ 
\hline
\begin{tabular}[c]{@{}c@{}}BatchNorm -- ReLU -- ConvLSTM2D -- Dropout -- BatchNorm -- ReLU -- ConvLSTM2D -- Dropout \\-- BatchNorm -- ConvLSTM2D -- ReLU -- BatchNorm -- ConvLSTM2D -- \textbf{Add}\\ \end{tabular} \\ 
\hline
\begin{tabular}[c]{@{}c@{}}BatchNorm -- ReLU -- ConvLSTM2D -- Dropout -- BatchNorm -- ReLU -- ConvLSTM2D -- Dropout \\-- BatchNorm -- ReLU -- ConvLSTM2D -- \textbf{Add}\\ \end{tabular} \\ 
\hline
\begin{tabular}[c]{@{}c@{}}BatchNorm -- ReLU -- ConvLSTM2D -- Dropout -- BatchNorm -- ReLU -- ConvLSTM2D -- Dropout \\-- BatchNorm -- ReLU -- ConvLSTM2D -- \textbf{Add}\\ \end{tabular} \\ 
\hline
\begin{tabular}[c]{@{}c@{}}BatchNorm -- ReLU -- ConvLSTM2D -- Dropout -- BatchNorm -- ReLU -- ConvLSTM2D -- Dropout \\-- BatchNorm -- ReLU -- ConvLSTM2D -- \textbf{Add}\\ \end{tabular} \\ 
\hline
\begin{tabular}[c]{@{}c@{}}BatchNorm -- ReLU -- ConvLSTM2D -- Dropout -- BatchNorm -- ReLU -- ConvLSTM2D -- Dropout \\-- BatchNorm -- ConvLSTM2D -- ReLU -- BatchNorm -- ConvLSTM2D -- \textbf{Add}\\ \end{tabular} \\ 
\hline
\begin{tabular}[c]{@{}c@{}}BatchNorm -- ReLU -- ConvLSTM2D -- Dropout -- BatchNorm -- ReLU -- ConvLSTM2D~-- Dropout \\-- BatchNorm -- ReLU -- ConvLSTM2D -- \textbf{Add}\\ \end{tabular} \\ 
\hline
\begin{tabular}[c]{@{}c@{}}BatchNorm -- ReLU -- ConvLSTM2D -- Dropout -- BatchNorm -- ReLU -- ConvLSTM2D -- Dropout \\-- BatchNorm -- ReLU -- ConvLSTM2D -- \textbf{Add}\\ \end{tabular} \\ 
\hline
\begin{tabular}[c]{@{}c@{}}BatchNorm -- ReLU -- ConvLSTM2D -- Dropout -- BatchNorm -- ReLU -- ConvLSTM2D -- Dropout \\-- BatchNorm -- ReLU -- ConvLSTM2D -- \textbf{Add}\\ \end{tabular} \\ 
\hline
\begin{tabular}[c]{@{}c@{}}BatchNorm -- ReLU -- ConvLSTM2D -- Dropout -- BatchNorm -- ReLU -- ConvLSTM2D -- Dropout \\-- BatchNorm -- ReLU -- ConvLSTM2D -- \textbf{Add}\\ \end{tabular} \\ 
\hline
\begin{tabular}[c]{@{}c@{}}BatchNorm -- ReLU -- ConvLSTM2D -- Dropout -- BatchNorm -- ReLU -- ConvLSTM2D -- Dropout \\-- BatchNorm -- ReLU -- ConvLSTM2D -- \textbf{Add}\\ \end{tabular} \\ 
\hline
\begin{tabular}[c]{@{}c@{}}BatchNorm -- ReLU -- ConvLSTM2D -- Dropout -- BatchNorm -- ReLU -- ConvLSTM2D -- Dropout \\-- BatchNorm -- ConvLSTM2D -- ReLU -- BatchNorm -- ConvLSTM2D -- \textbf{Add}\\ \end{tabular} \\ 
\hline
\begin{tabular}[c]{@{}c@{}}BatchNorm -- ReLU -- ConvLSTM2D -- Dropout -- BatchNorm -- ReLU -- ConvLSTM2D -- Dropout \\-- BatchNorm -- ReLU -- ConvLSTM2D -- \textbf{Add}\\ \end{tabular} \\ 
\hline
\begin{tabular}[c]{@{}c@{}}BatchNorm -- ReLU -- ConvLSTM2D -- Dropout -- BatchNorm -- ReLU -- ConvLSTM2D -- Dropout \\-- BatchNorm -- ReLU -- ConvLSTM2D -- \textbf{Add}\\ \end{tabular} \\ 
\hline
BatchNorm -- GlobalAveragePooling3D -- Dropout -- Dense \\
\hline
\end{tabular}
}
\end{table*}
\end{document}